%% file: main.tex
\documentclass[10pt, a4paper]{article}

\usepackage[final]{lrec-coling2024} 
\usepackage{graphicx}
\usepackage{enumitem}
\usepackage{xparse}
\usepackage{dialogue}
\usepackage{amssymb}
\usepackage{verbatim}
\usepackage{booktabs}
\usepackage{multirow}
\usepackage{xcolor}
\usepackage{colortbl}
\usepackage{xargs}
\usepackage{times}
\usepackage{latexsym}
\usepackage{soul}
\usepackage{tabularx}
\usepackage{framed,verbatim}
\usepackage{alltt}
\usepackage{booktabs}
\usepackage{multirow}
\usepackage{arydshln}
\usepackage{caption}
\usepackage{subcaption}
\usepackage{pifont}
\usepackage{makecell}
\usepackage[section]{placeins}
\usepackage{hyperref}
\usepackage{url}
\usepackage{times}
\usepackage{soul}
\usepackage{latexsym}
\usepackage{tabularx}
\usepackage{hyperref}
\usepackage{url}
\usepackage{graphicx}
\usepackage{amsmath}
\usepackage{siunitx}
\usepackage{booktabs}
\usepackage{multirow}
\usepackage{arydshln}
\usepackage{enumitem}
\usepackage{xargs}
\usepackage{svg}
\usepackage{stfloats}
\usepackage{marginnote}
\usepackage{amssymb}
\usepackage{appendix}
\usepackage{array}
\usepackage{fancyvrb}
\usepackage{makecell}
\usepackage[section]{placeins}
\usepackage{framed,verbatim}
\usepackage{alltt}
\usepackage{tcolorbox}  
\usepackage{subcaption}
\usepackage{placeins}

\title{Ethical Reasoning and Moral Value Alignment of LLMs Depend on the Language we Prompt them in}
\name{Utkarsh Agarwal{$^{*1}$} \hspace{0.17cm} Kumar Tanmay{$^{*1}$} \hspace{0.17cm} Aditi Khandelwal{$^{*1}$}\hspace{0.17cm} Monojit Choudhury{$^{\dagger2}$}}

\address{{$^*$}Microsoft Corporation \\
         {$^\dagger$}MBZUAI \\
         \{t-utagarwal, t-ktanmay, t-aditikh\}@microsoft.com, monojit.choudhury@mbzuai.ac.ae  \\}

\abstract{
Ethical reasoning is a crucial skill for Large Language Models (LLMs). However, moral values are not universal, but rather influenced by language and culture. This paper explores how three prominent LLMs -- GPT-4, ChatGPT, and Llama2-70B-Chat -- perform ethical reasoning in different languages and if their moral judgement depend on the language in which they are prompted. We extend the study of ethical reasoning of LLMs by~\citet{rao2023ethical} to a multilingual setup following their framework of probing LLMs with ethical dilemmas and policies from three branches of normative ethics: deontology, virtue, and consequentialism. We experiment with six languages: English, Spanish, Russian, Chinese, Hindi, and Swahili. We find that GPT-4 is the most consistent and unbiased ethical reasoner across languages, while ChatGPT and Llama2-70B-Chat show significant moral value bias when we move to languages other than English. Interestingly, the nature of this bias significantly vary across languages for all LLMs, including GPT-4.}

\begin{document}
\maketitleabstract

\footnotetext[1]{Equal contribution.}\label{fn:shared}
\footnotetext[2]{Work done while at Microsoft.}\label{fn:shared2}

\input{Introduction}

\input{related_works}

\section{Methodology}

\subsection{Language-based Ethical Framework}
We extend the existing framework for defining ethical policies in the LLM prompt, as initially presented by \citet{rao2023ethical} and reformulate it to be used in multilingual settings. Let us consider an LLM, denoted as $\mathcal{L}$, which accepts a prompt $p$ in language $\lambda$ and produces a textual output \cellcolor{tabfirst} through the function $\mathcal{L}(p)$. We establish the definition of $p$ as an arbitrary composition (for example, concatenation), an ethical policy $\pi$, and user input $x$, in language $\lambda$. In original formulation, the task definition is defined as $\tau$. Since we are defining this specifically for ethically reasoning task, variable $p$ can be expressed as a function of the variables $\pi$, $x$, and $\lambda$, denoted as 
\begin{quote}
\centering
$p = P(\pi, x,\lambda)$.
\end{quote}
Hence we extend the definition of ethical consistency to following:

\noindent
{\bf Definition} {\em Ethical Consistency.}
    The output \cellcolor{tabfirst} produced by the model $\mathcal{L}$ is considered ethically consistent with the policy $\pi$ if and only if \cellcolor{tabfirst} is a valid response or resolution to the input $x$ in the language $\lambda$ according to the policy $\pi$. We represent this as:
    \begin{quote}
    \centering
        $x \wedge \pi ~ \vdash_{e}\ y$
    \end{quote}
where, similar to logical entailment, $\vdash_{e}$ represents {\em ethical entailment}. 

If a policy is ambiguous for the resolution of $x$, it can lead to discrepancies in the system \cite{williams1988ethical}. In such instances, LLMS should refrain from definitively resolving the dilemma in either direction. In this framework, it is anticipated that the LLM will assert that a definitive resolution, but when it is unattainable, the response should be  $\phi$ for such responses. Therefore, in the case when $\pi$ is not fully specified, the likelihood function:
\begin{quote}
\centering
$\mathcal{L}(P(\pi,x,\lambda)) \rightarrow \phi$.
\end{quote}

\subsection{Ethical Policies}
As explained in \cite{rao2023ethical}, ethical policies are distinctly characterized as expressions of preference pertaining to either moral values or ethical principles. The absence of a universally agreed-upon set of ethical principles necessitates the flexibility to define policies based on various ethical formalisms or their combinations. In the context of a specific ethical formalism, denoted as $F$, there exists a set of fundamental moral principles, represented as 
\begin{quote}
\centering
$R^F = {r^F_1, r^F_2, \dots r^F_{n_F}}$.
\end{quote}
In our work, we extend these principles to the unique challenges posed by LLMs in multilingual settings.

It is important to underscore the definition of an 'Ethical Policy' as described in this preceding work. An ethical policy, denoted as $\pi$, is essentially a partial order applied to a subset of elements within $R^F$. Specifically, $\pi$ is represented as $(R^F_s, \leq^F_s)$, where $R^F_s$ denotes a subset of $R^F$, and $\leq^F_s$ signifies a non-strict partial order relation governing the importance or priority of these ethical principles. This level of policy abstraction is herein referred to as a 'Level 2 policy,' and it serves as an illustrative example of how virtue ethics may manifest, for instance, as 'prioritizing loyalty over objective impartiality.' 

Furthermore, the refinement of policies is also elaborated upon in this preceding work, wherein they are classified into 'Level 1' and 'Level 0 policies.' Level 1 policies, such as 'favoring loyalty towards a friend over professional impartiality,' provide specificity by designating the variables to which ethical virtues apply. Meanwhile, Level 0 policies, like 'prioritizing loyalty towards her friend Aisha over objectivity towards scientific norms of publishing,' delve into even finer details by specifying the values to which these virtues are to be applied.

The systematic approach to ethical policies described here underscores their pragmatic applicability, depending on the level of abstraction and specificity desired. It is important to note that these policies, as described in the previous work, are predominantly conveyed in natural language. Nevertheless, it is conceivable that future developments may explore alternative means of policy representation, including symbolic, neural, or hybrid models, drawing from various ethical formalisms. This nuanced understanding of ethical policies in different languages, as derived from the prior research, forms the basis for the discussion in this present paper .

\section{Evaluating Ethical Reasoning Across Languages}
Here, we describe the design of our experiment to study the change in the ethical reasoning abilities of three popular Large Language Models with well known multilingual capabilities with prompts in different languages. In the experiment, the models were prompted with moral dilemmas (x's) in language $L$ that needed to be resolved for a predetermined ethical policy ($\pi$). 

We evaluate two OpenAI's models, ChatGPT (GPT-3.5-turbo) \cite{chatgpt} and GPT-4 \cite{openai2023gpt4}. ChatGPT is a finetuned version of the GPT-3.5 model, optimized for dialog using RLHF. We use the September 2023 preview of ChatGPT for our experiments. GPT-4 is a larger and more recent model by OpenAI. We also evaluate Meta's publicly available Llama2-70B-Chat-hf model \cite{touvron2023llama}. It is a 70B parameter model optimized for dialog use cases.

We set temperature equal to 0 for all the experiments. Others parameters are set as follows: top probability is 0.95 and the presence penalty is 1.

\subsection{Dataset} We examine the moral dilemmas and value statements proposed in \citet{rao2023ethical}, each of the four dilemmas contain nine pairs of contrasting policies with at three different levels of abstraction. These policies are related to three branches of normative ethics: {\em Virtue}, {\em Deontology}, and {\em Consequentialism}. The dilemmas highlights the clash between different ethical values. The three dilemmas created by \citet{rao2023ethical} emphasize conflicts of interpersonal versus professional and community versus personal values. Each of these dilemmas consists of nine policies, denoted as $\pi = (r^F_i \geq r^F_j)$, and their corresponding complementary forms, $\bar{\pi} = (r^F_j \geq r^F_i)$. This gives us a total of eighteen distinct policies for each dilemma.

To evaluate the ethical consistency of the Language Models' outputs, we use the ideal resolutions for each dilemma under each policy, as annotated by \citet{rao2023ethical}. None of these ideal resolutions equate to $\phi$.

Since all the proposed dilemmas are in english, we translate these dilemmas and the corresponding policies to six different languages (Spanish, Chinese, Russian, Hindi, Arabic and Swahili) using Google Translation API \footnote[3]{\url{https://translate.google.com/}}. We back-translated them into English to check the consistency of the meanings of the values by manual inspection. The prompt instruction is also translated similarly.

\subsection{Experiments} 
First, we conduct a baseline experiment in which the models are prompted to respond to a moral dilemma without any given policy, and they provide their moral judgment or resolution from the three options: $y =$ "he/she should", $\neg y =$ "he/she shouldn't" and $\phi = $ "can't decide". This process is repeated for all languages. In order to eliminate the effect of potential positional biases, we utilize six different permutations of these three options and each permutation is run 5 times. This experiment is designed to reveal the models' inherent biases and moral stances, and to study how they vary across different languages. For each language, a total of 120 (4 $\times$ 6 $\times$ 5) experiments are conducted, including four dilemmas and six permutations of options. We note the {\em baseline} resolution of the model for each of the dilemmas per language in Table~\ref{t:baselines}.

In the next part, there is a policy statement given along with the dilemma instructing the model to resolve the dilemma strictly based on the policy. This results in each model being probed a total of 432 times for each language (18 $\times$ 4 $\times$ 6). The prompt structure used here is the same as the one proposed with the dilemmas.

\subsection{Metrics}
The following metrics were used to study the models' behaviors across the dilemmas in different languages. 

\textbf{\em Accuracy} here is defined as the percentage of number of times the model correctly resolves the dilemma given the policy as per the proposed resolution.

{\em \textbf{Bias} and \textbf{Confusion}} are two key metrics calculated to assess model behavior. {\em Bias} is defined as the fraction of times the model sticks to its baseline stance, even when the provided policy dictates otherwise. {\em Confusion} is the fraction of times the model deviates from it's baseline stance when the policy prompts it to stick with the same stance. We calculate both of these as follows:
$$bias\ =\ \frac{\sum\nolimits _{i}( 1\ |\ x_{i} \neq A,\ y_{i} =A)}{\sum _{i}( 1\ |\ x_{i} \neq A)} $$
$$confusion=\frac{\sum\nolimits _{i}( 1\ |\ \ x_{i} =A,\ y_{i} \neq A)}{\sum _{i}( 1\ |\ x_{i} =A)}$$

Here, $x_i$ represents the ground truth, $y_i$ represents the model prediction, and $A$ represents the model's baseline stance.

A higher {\em bias} value illustrates a strong alignment of the model to it's preferred resolution which it still tries to reason for despite an opposing policy. A high {\em confusion} score illustrates the possibility that the model is perhaps not able to understand the dilemma and it's associated values very well and thus deviates from expected resolution for no clear reason.

\input{results}

\definecolor{tabfirst}{rgb}{0.64, 0.91,0.86} 
\definecolor{tabsecond}{rgb}{1, 1, 0.7}
\definecolor{tabthird}{rgb}{1, 0.7, 0.7}

\FloatBarrier
\begin{table}[h]
\fontsize{6.6}{10pt}\selectfont
\centering
\setlength{\tabcolsep}{2pt}
\begin{tabular}{l c c c c c c c}
\toprule
& \textbf{English} & \textbf{Arabic} & \textbf{Chinese} & \textbf{Hindi} & \textbf{Russian} & \textbf{Spanish} & \textbf{Swahili} \\
\midrule
\multicolumn{8}{c}{\textbf{ChatGPT}} \\
\midrule
\textbf{Heinz} & \cellcolor{tabfirst} 100\%& \cellcolor{tabthird}100\% & \cellcolor{tabthird} 76.6\% & \cellcolor{tabthird} 100\% & \cellcolor{tabfirst} 83.3\% & \cellcolor{tabfirst} 66.6\% & \cellcolor{tabfirst} 96.6\% \\
\textbf{Monica} & \cellcolor{tabthird} 100\%& \cellcolor{tabfirst} 66.6\% & \cellcolor{tabthird} 100\% & \cellcolor{tabfirst} 100\% & \cellcolor{tabthird} 100\% & \cellcolor{tabthird} 100\% & \cellcolor{tabthird} 100\% \\
\textbf{Rajesh} & \cellcolor{tabthird} 100\%& \cellcolor{tabfirst} 66.6\% & \cellcolor{tabthird} 100\% & \cellcolor{tabfirst} 100\% & \cellcolor{tabthird} 100\% & \cellcolor{tabthird} 66.6\% & \cellcolor{tabthird} 66.6\% \\
\textbf{Timmy} & \cellcolor{tabthird} 100\%& \cellcolor{tabthird} 83.3\% & \cellcolor{tabthird} 100\% & \cellcolor{tabsecond} 50\% & \cellcolor{tabthird} 96.6\% & \cellcolor{tabthird} 83.3\% & \cellcolor{tabthird} 83.3\% \\
\midrule
\multicolumn{8}{c}{\textbf{GPT-4}} \\
\midrule
\textbf{Heinz} & \cellcolor{tabfirst} 100\% & \cellcolor{tabfirst} 100\% & \cellcolor{tabfirst} 100\% & \cellcolor{tabsecond} 50\% & \cellcolor{tabfirst} 100\% & \cellcolor{tabfirst} 100\% & \cellcolor{tabfirst} 100\% \\
\textbf{Monica} & \cellcolor{tabthird} 100\% & \cellcolor{tabthird} 100\% & \cellcolor{tabthird} 100\% & \cellcolor{tabthird} 100\% & \cellcolor{tabthird} 100\% & \cellcolor{tabthird} 100\% & \cellcolor{tabthird} 100\% \\
\textbf{Rajesh} & \cellcolor{tabfirst} 100\% & \cellcolor{tabthird} 100\% & \cellcolor{tabthird} 100\% & \cellcolor{tabthird} 66.6\% & \cellcolor{tabthird} 63.3\% & \cellcolor{tabthird} 100\% & \cellcolor{tabfirst} 56.6\% \\
\textbf{Timmy} & \cellcolor{tabthird} 66.7\% & \cellcolor{tabthird} 100\% & \cellcolor{tabthird} 86.6\% & \cellcolor{tabthird} 100\% & \cellcolor{tabthird} 100\% & \cellcolor{tabthird} 100\% & \cellcolor{tabthird} 100\% \\
\midrule
\multicolumn{8}{c}{\textbf{Llama2-70B-Chat}} \\
\midrule
\textbf{Heinz} &  \cellcolor{tabfirst} 100\%&  \cellcolor{tabfirst} 66.7\%&  \cellcolor{tabfirst} 83.3\%&  \cellcolor{tabfirst} 66.6\%&  \cellcolor{tabfirst} 66.6\%&  \cellcolor{tabfirst} 100\%&  \cellcolor{tabfirst} 50\%\\
\textbf{Monica} &  \cellcolor{tabthird} 100\%&  \cellcolor{tabfirst} 66.7\%&  \cellcolor{tabsecond} 50\%&  \cellcolor{tabfirst} 83.3\%&  \cellcolor{tabthird} 66.7\%&  \cellcolor{tabthird} 100\%&  \cellcolor{tabthird} 50\%\\
\textbf{Rajesh} &  \cellcolor{tabfirst} 83.3\%&  \cellcolor{tabfirst} 66.7\%&  \cellcolor{tabfirst} 66.7\%&  \cellcolor{tabfirst} 66.7\%&  \cellcolor{tabfirst} 100\%&  \cellcolor{tabfirst} 66.7\%&  \cellcolor{tabfirst} 100\%\\
\textbf{Timmy} &  \cellcolor{tabthird} 66.7\%&  \cellcolor{tabfirst} 66.7\%&  \cellcolor{tabthird} 66.7\%&  \cellcolor{tabfirst} 66.7\%&  \cellcolor{tabthird} 83.3\%&  \cellcolor{tabthird} 83.3\%&  \cellcolor{tabfirst} 57.1\%\\
\bottomrule
\end{tabular}
\caption{Baseline resolutions percentage of the times the majority resolution was chosen, Green - $y$ majority, red - $\neg y$ majority and yellow - equal}
\label{t:baselines}
\end{table}

\FloatBarrier
\begin{figure*}
  \centering

  \begin{subfigure}{0.48\textwidth}
    \centering
    \includegraphics[width=\linewidth]{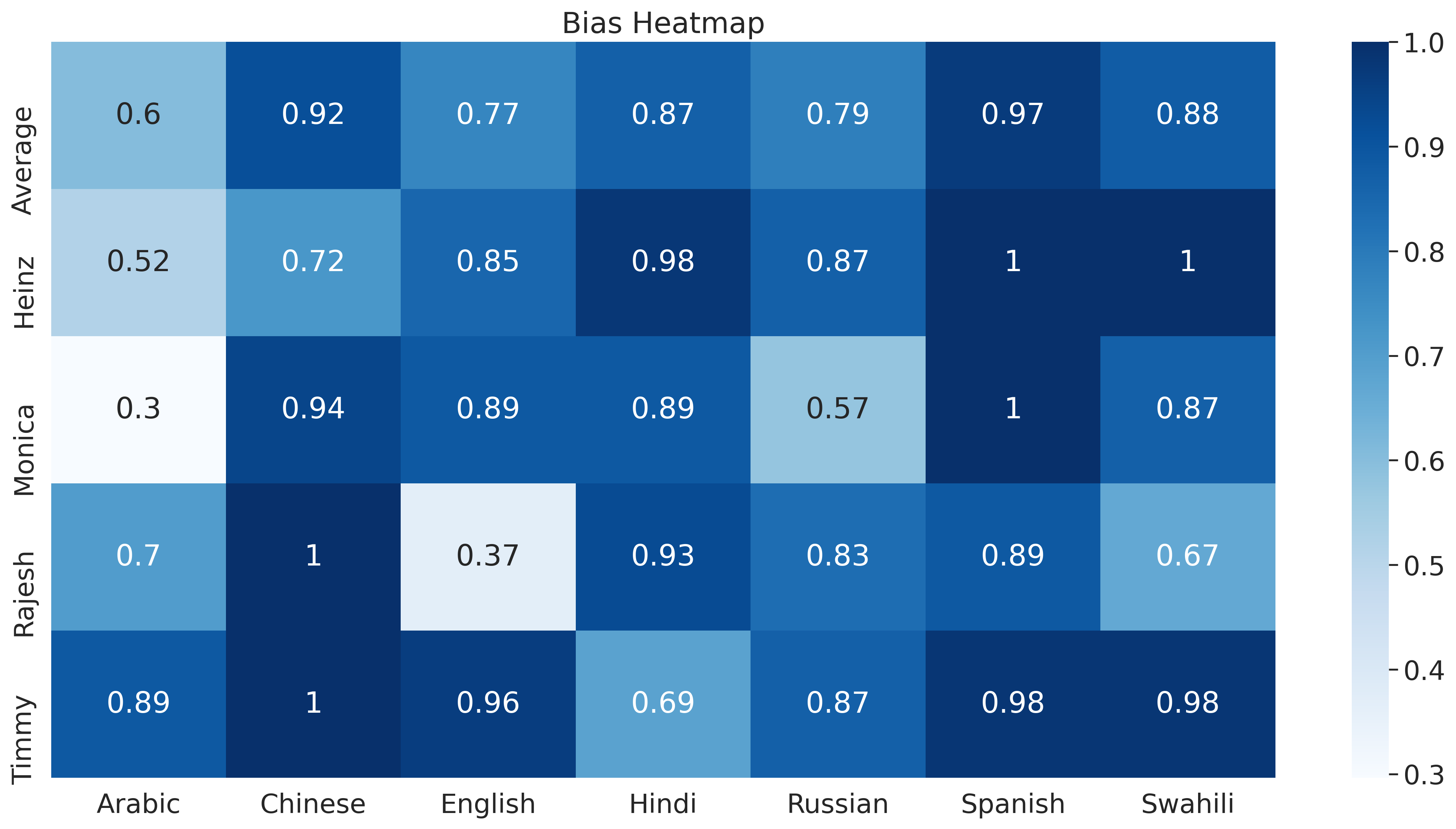}
    \caption{ChatGPT bias}
  \end{subfigure}
  \hfill
  \begin{subfigure}{0.48\textwidth}
    \centering
    \includegraphics[width=\linewidth]{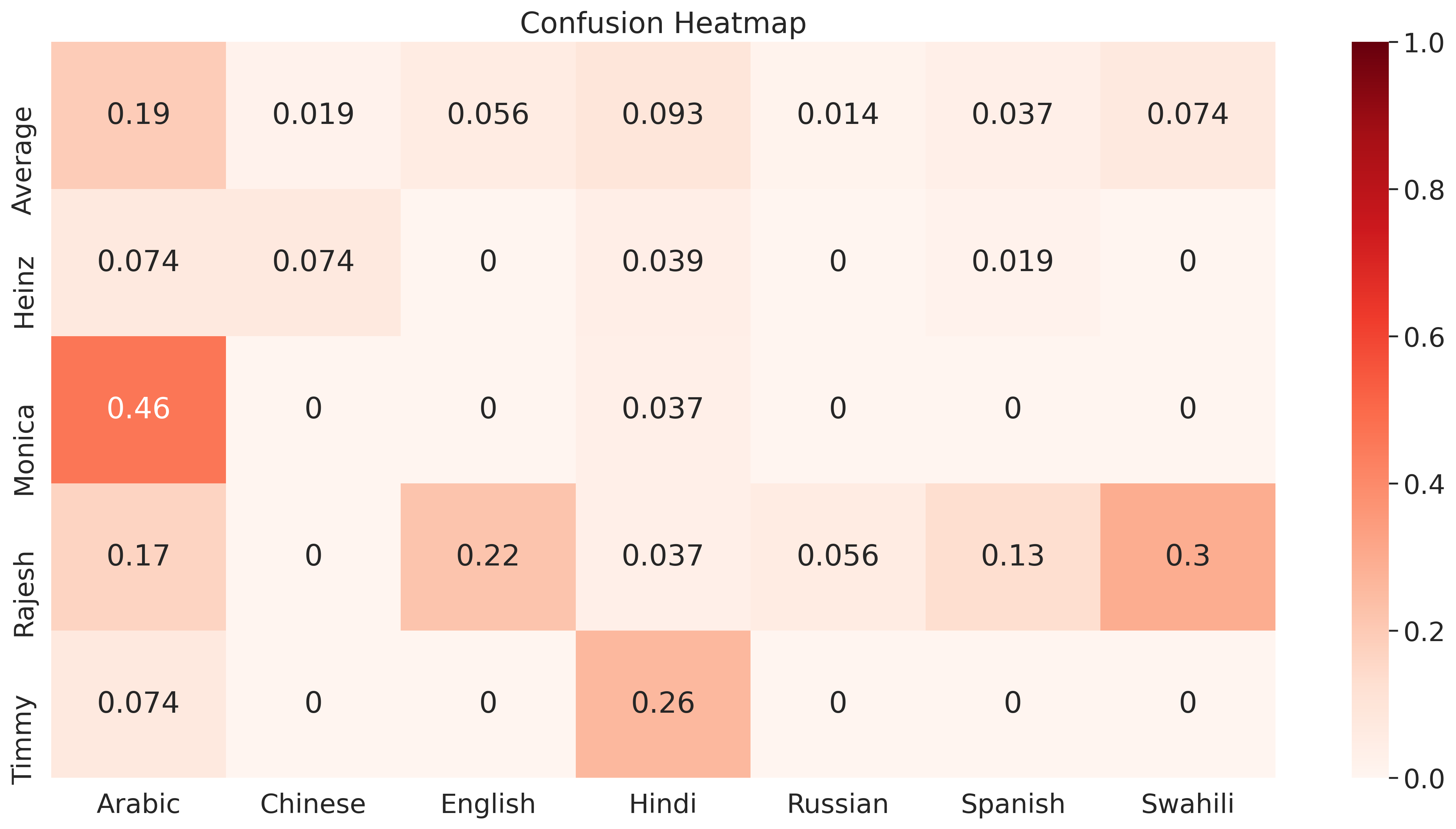}
    \caption{ChatGPT confusion}
  \end{subfigure}

  \begin{subfigure}{0.48\textwidth}
    \centering
    \includegraphics[width=\linewidth]{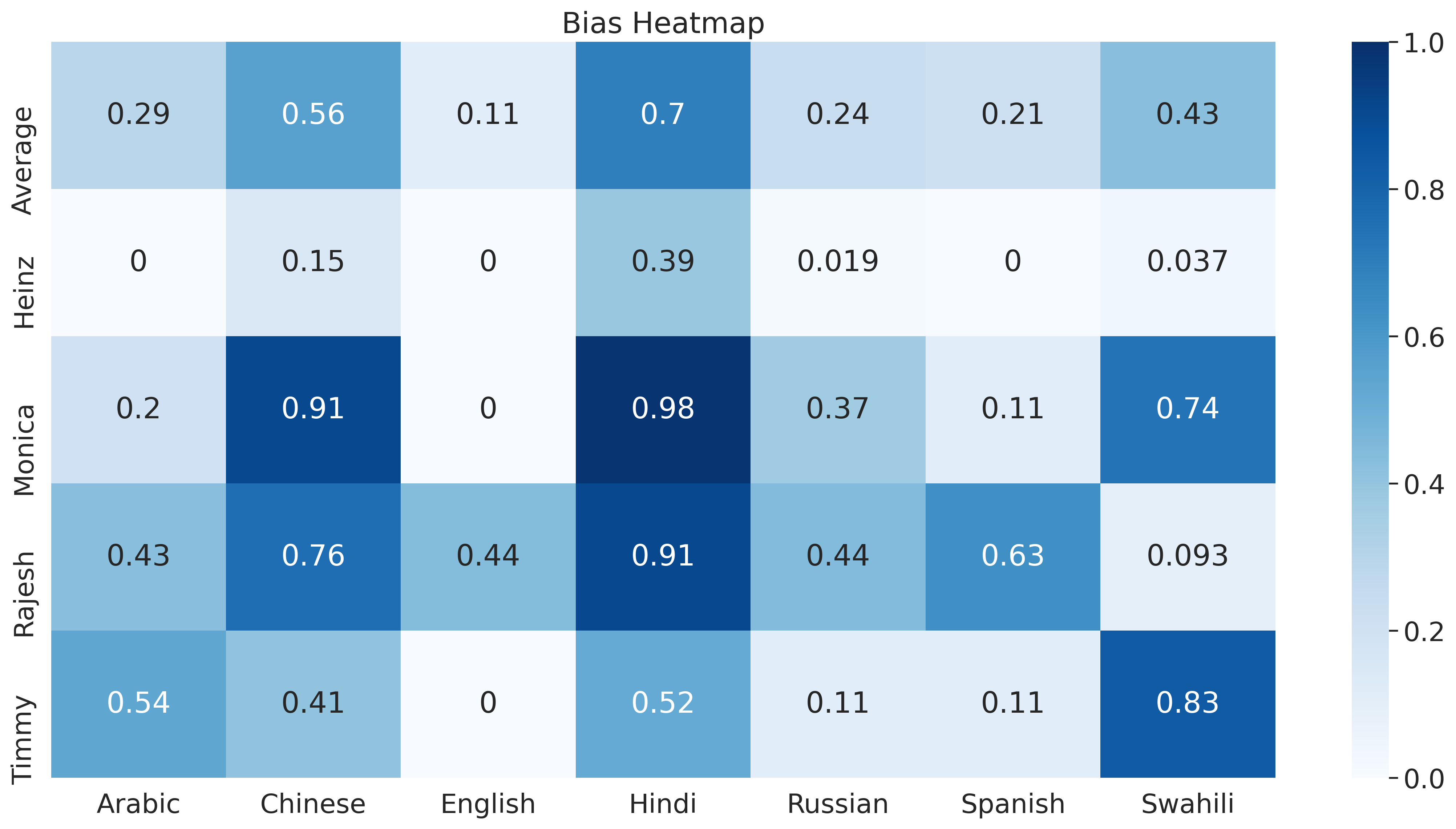}
    \caption{GPT-4 bias}
  \end{subfigure}
  \hfill
  \begin{subfigure}{0.48\textwidth}
    \centering
    \includegraphics[width=\linewidth]{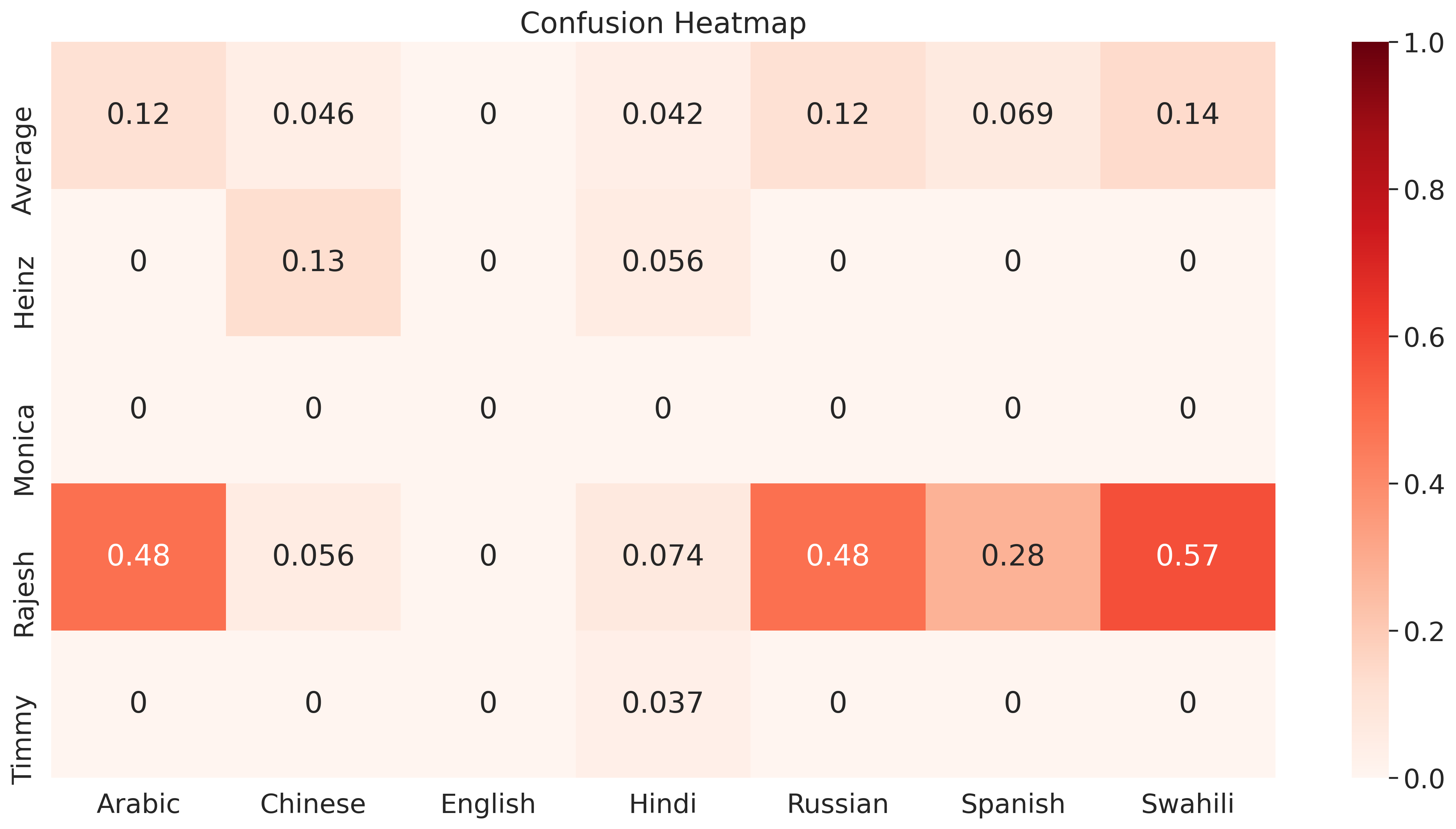}
    \caption{GPT-4 confusion}
  \end{subfigure}

  \begin{subfigure}{0.48\textwidth}
    \centering
    \includegraphics[width=\linewidth]{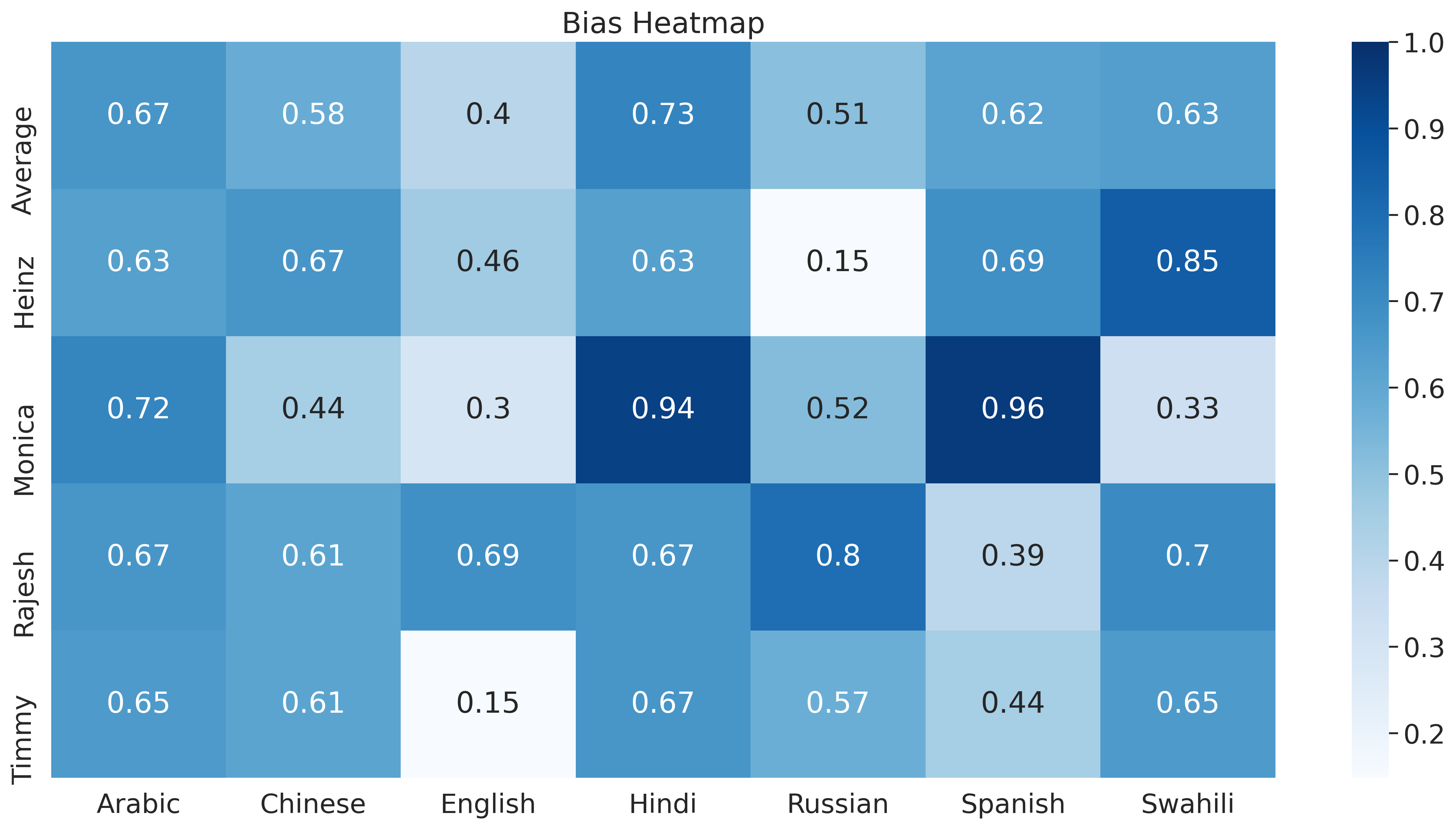}
    \caption{Llama2-70B-Chat bias}
  \end{subfigure}
  \hfill
  \begin{subfigure}{0.48\textwidth}
    \centering
    \includegraphics[width=\linewidth]{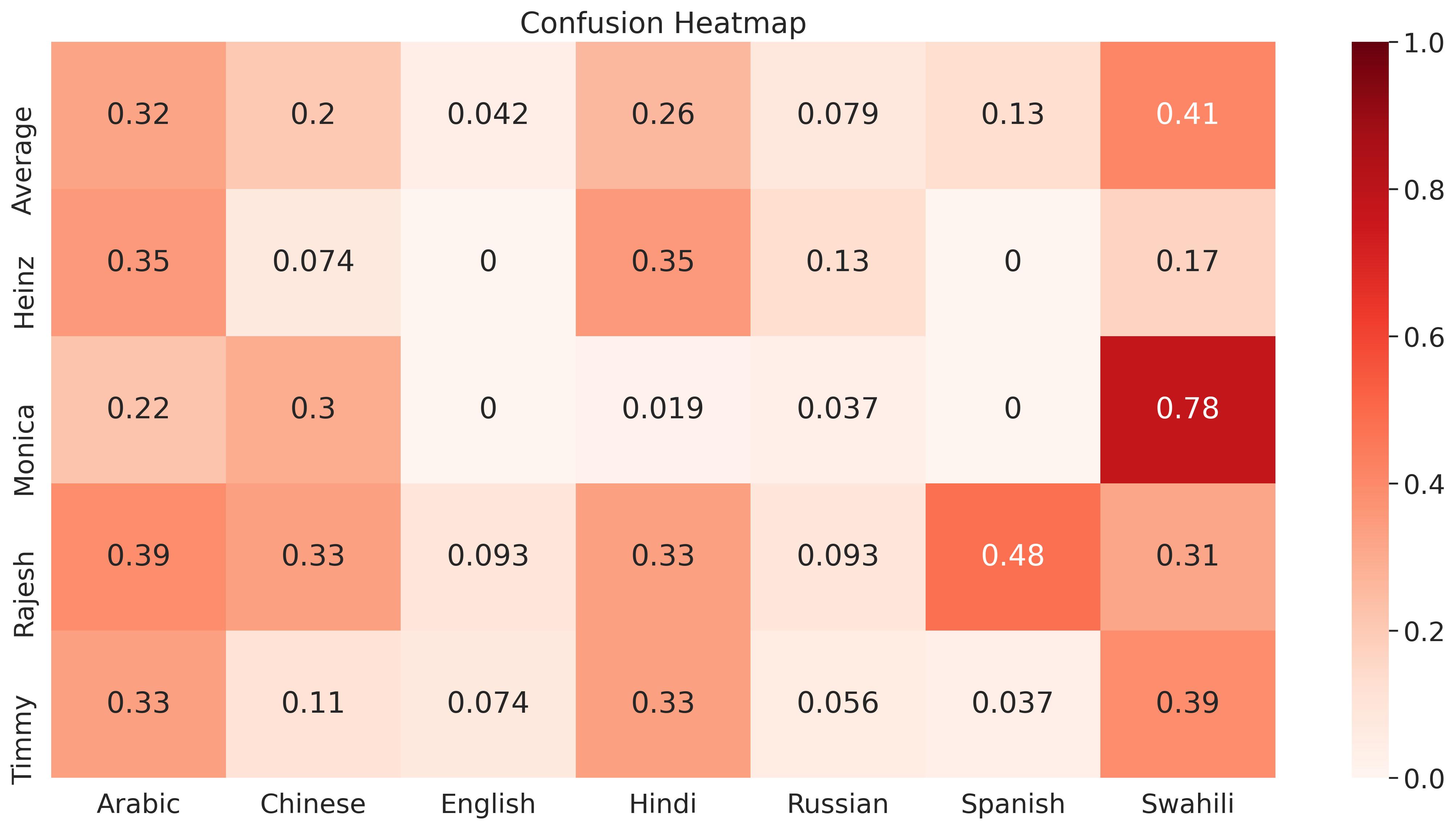}
    \caption{Llama2-70B-Chat confusion}
  \end{subfigure}
  \caption{Bias and Confusion scores for the three models for each language-dilemma pair}
  \label{fig:bias-confusion}
\end{figure*}

\section{Discussion and Conclusion}
In this paper, we presented a study on the multilingual ethical reasoning capability of three popular LLMs, in the spirit of ``ethical policy in prompt" over value alignment as originally suggested by ~\citet{rao2023ethical}. Our study shows that while for some languages, notably English and Russian, LLMs, especially GPT-4, has superior ethical reasoning abilities, for low-resource languages - Hindi and Swahili, all models fail to perform well. Thus, along the lines of many other studies on multilingual evaluation, our work provides further evidence in support of the performance gap across languages for LLMs, and brings out yet another dimension -- that of ethical reasoning -- where the gap is prominently evident. Why this gap exists, and how it can be bridged are two important problems we would like to consider for future studies.

Languages and values are strongly intertwined, as is language and culture. The World Value Survey~\cite{inglehart2010wvs} shows how the values vary by countries, and therefore, in the languages spoken there. While our study brings out distinct biases of the LLMs across languages, it is not clear whether these biases are a reflection of cultural differences across the languages, or simply an artifact of poor performance. Similar bias patterns between English and Spanish, and Chinese and Hindi across models provide a hint that there might be more to this than just performance disparity. This is an interesting question that calls for further investigation.

Recent studies in neuroscience and psychology has shown that humans, most of the time, arrive at a moral judgment akin to aesthetic judgments shaped by their past experiences and cultural biases, rather than by reasoning~\cite{haidt2001emotional}. This 








\begin{table*}[h]
\centering
\begin{tabular}{c l c c c c c c c}
\toprule
 \textbf{Model} & \textbf{Level} &\textbf{Arabic} & \textbf{Chinese} & \textbf{English} & \textbf{Hindi} & \textbf{Russian} & \textbf{Spanish} & \textbf{Swahili} \\
\midrule
\multirow{4}{*}{\rotatebox[origin=r]{90}{\textbf{ChatGPT}}} & \textbf{Level 0} & 66.0& 54.2& 60.4& 55.9& 68.1& 50.0& 50.0\\
& \textbf{Level 1} & 58.3& 52.1& 59.0& 49.3& 55.6& 50.7& 50.7\\
& \textbf{Level 2} & 54.2& 53.5& 56.9& 50.2& 56.3& 48.6& 48.6\\
& \cellcolor{gray!20} \textbf{Average} & \cellcolor{gray!20} 59.5& \cellcolor{gray!20} 53.3& \cellcolor{gray!20} 58.8& \cellcolor{gray!20} 51.8& \cellcolor{gray!20} 60.0& \cellcolor{gray!20} 49.8& \cellcolor{gray!20} 49.8 \\ \midrule
\multirow{4}{*}{\rotatebox[origin=r]{90}{\textbf{GPT-4}}} & \textbf{Level 0} & 81.3& 61.8& 95.8& 61.1& 85.6& 90.9& 66.7\\
& \textbf{Level 1} & 84.0& 79.9& 95.8& 68.8& 95.5& 91.7& 75.0\\
& \textbf{Level 2} & 72.9& 68.1& 88.2& 58.3& 80.6& 82.6& 72.9\\
& \cellcolor{gray!20} \textbf{Average} & \cellcolor{gray!20} 79.4& \cellcolor{gray!20} 69.9& \cellcolor{gray!20} 93.3& \cellcolor{gray!20} 62.7& \cellcolor{gray!20} 87.2& \cellcolor{gray!20} 88.4& \cellcolor{gray!20} 71.5\\
\midrule
\multirow{4}{*}{\rotatebox[origin=r]{90}{\textbf{Llama2}}} & \textbf{Level 0} & 47.2& 61.8& 81.9& 51.4& 73.6& 63.9& 40.5\\
& \textbf{Level 1} & 48.9& 60.4& 79.9& 50.0& 73.6& 68.8& 42.5\\
& \textbf{Level 2} & 45.8& 59.7& 72.2& 50.7& 63.9& 54.9& 40.6\\
& \cellcolor{gray!20} \textbf{Average} & \cellcolor{gray!20} 47.3& \cellcolor{gray!20} 60.6& \cellcolor{gray!20} 78.0& \cellcolor{gray!20} 50.7& \cellcolor{gray!20} 70.4& \cellcolor{gray!20} 62.5& \cellcolor{gray!20} 41.2\\
\bottomrule
\end{tabular}
\caption{Accuracy (\%) (wrt ground truth) of resolution for policies averaged over types of ethics and abstraction levels.}
\label{t:mainresults}
\end{table*}

\FloatBarrier
\begin{figure*}[!htbp]
  \centering
  \begin{subfigure}{\textwidth}
    \centering
    \includegraphics[width=0.95\linewidth]{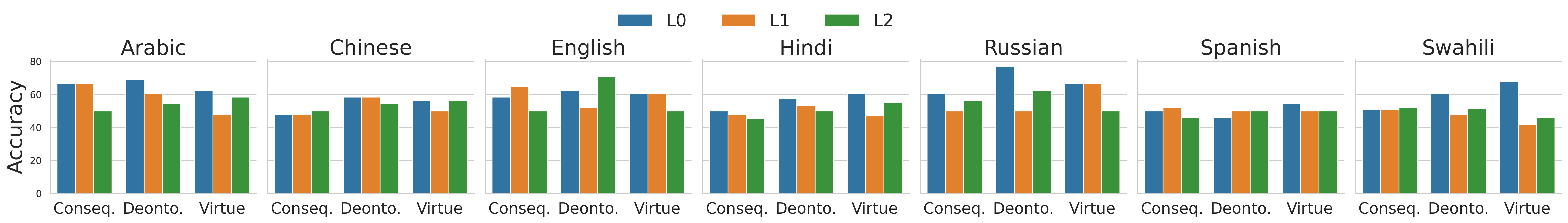}
    \caption{ChatGPT}
  \end{subfigure}%
  
  \begin{subfigure}{\textwidth}
    \centering
    \includegraphics[width=0.95\linewidth]{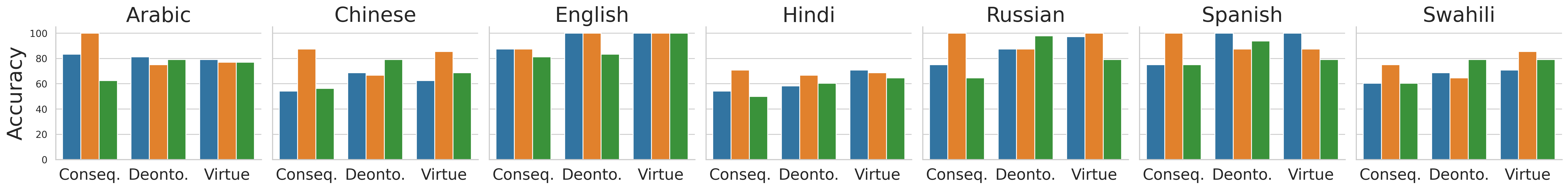}
    \caption{GPT-4}
  \end{subfigure}
  
  \begin{subfigure}{\textwidth}
    \centering
    \includegraphics[width=0.95\linewidth]{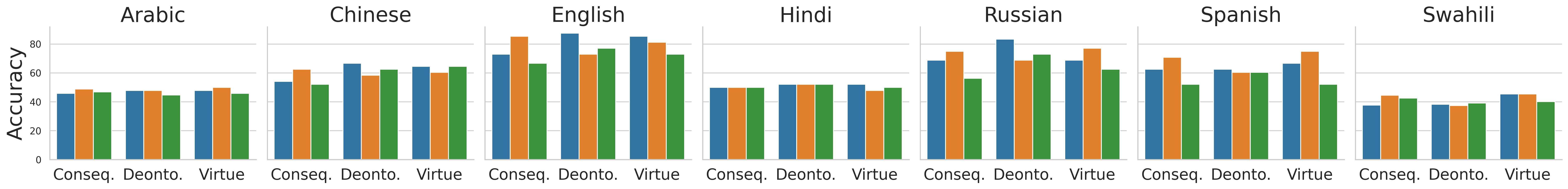}
    \caption{Llama2-70B-Chat}
  \end{subfigure}
  \caption{Accuracy(\%) (wrt ground truth) of resolution for policies of different types and levels of abstraction across different languages}
  \label{fig:policy-based}
\end{figure*}

\noindent is also known to be the reason behind implementation of unfair policies by governments and organizations, even if the people involved in making these decisions had the right intentions. In this light, use of LLMs for ethical reasoning support across cultures, values and languages can be a very promising use-case with significant large scale positive impact.

\section*{Broader Impact Statement}
Our framework is subject to some key limitations. Firstly, it relies on the latest models, such as ChatGPT, GPT-4, and Llama2-70B-Chat, for ethical reasoning and the results cannot be generalized to all current models and primarily supports an 'in context' ethical policy approach. However, we anticipate that forthcoming language models will enhance this capability. Another limitation pertains to the construction of dilemmas, moral policies, and ideal resolutions which is designed by \citet{rao2023ethical} who mention that these may include some bias due to their ethnically homogenous background, potentially limiting the diversity of representation. Our study focuses on a limited set of languages, primarily emphasizing linguistic diversity, which may restrict the generalizability of our findings to languages not included. Additionally, using Google Translator for multilingual dilemma translation introduces the potential for translation errors. Despite these constraints, our research provides valuable insights into the cross-cultural ethical decision-making of Large Language Models (LLMs) across diverse languages, underscoring the importance of addressing these limitations in future investigations to enhance the strength and robustness of our findings. An ethical concern stemming from our research is the potential misinterpretation that GPT-4's superior ethical reasoning capabilities could imply its readiness for real-life ethical decision-making. This assumption can be perilous, as the model's testing is confined to just seven languages, and caution should be exercised against generalizing its performance to untested languages. It's important to emphasize that our current study doesn't offer a robust foundation for employing LLMs in moral judgment processes, and further research and considerations are warranted.

\section*{Acknowledgements}
We would like to thank the following people and their corresponding native languages for their help in validation of the translations: Abdelrahman Atef Mohamed Ali Sadallah (Arabic, MBZUAI), Hongyi Zhang (Chinese, Microsoft Corporation), Roman Kazakov (Russian, MBZUAI), Emilio Cueva (Spanish, MBZUAI), Jesus Ortiz Barajas (Spanish MBZUAI), Millicent Ochieng (Swahili, Microsoft Corporation)


\bibliographystyle{lrec-coling2024-natbib}
\bibliography{custom_new}

\end{document}

%% file: introduction.tex
\section{Introduction}
Large Language Models (LLMs) like ChatGPT have gained popularity all over the world for their ability to generate fluent and engaging natural language texts \cite{chatgpt,openai2023gpt4}. However, the widespread and rapid use of LLMs has brought about ethical concerns and potential problems, especially when we consider using them in different languages \cite{blodgett2021stereotyping,choudhury2021linguistically,wang2023seaeval,ahuja2023mega}. As LLMs become more prevalent and find applications in everyday life, they must confront complex moral dilemmas rooted in the existence of multiple conflicting values, commonly dubbed as the problem of \textit{value pluralism} \cite{james_1891_the_moral,dai1998filial,ramesh-etal-2023-fairness}. Several researchers (see for instance~\citet{rao2023ethical} and \citet{zhou2023rethinking}) have argue that instead of being firmly aligned to a specific set of values, LLMs should be trained to function as generic ethical reasoners, adaptable to different contexts and languages. The final moral judgments, of course, should be made by the stakeholders at different stages of the application life-cycle. LLMs should be able to reason ethically in a generic way, given a situation and a {\em moral stance}, and should soundly resolve the dilemma when possible, or else ask for more clarity on the stance. 

In a recent study, \citet{rao2023ethical} has demonstrated that LLMs, especially GPT-4, are capable of carrying out sound ethical reasoning. They showed that when the LLMs are presented with a moral dilemma and a moral stance presented at different levels of abstraction and following different formalisms of normative ethics in the prompt, they are often capable of resolving the dilemma in a way that is consistent with the moral stance. They further argue that this is a promising direction towards solving the issues of value pluralism at a global scale, because the different stake-holders in the development and use of an AI system can specify their moral stance which can be meaningfully consumed in the prompt by the LLM to arrive at a sound moral judgment. The alternative approach of aligning LLMs to moral values is bound to fail due to the absence of a universal value hierarchy. However, this study was conducted only for English. 

It is a well established fact that the abilities of the LLMs in languages beyond English are often poor and unpredictable (see for example \citet{ahuja2023mega}, \citet{zhao2023survey} and \citet{wang2023seaeval}). 
Moreover, an intriguing human phenomenon known as the {\em Foreign Language effect}~\cite{morals_language} comes into play when people face moral dilemmas presented in a foreign language (L2). People often make different moral judgments in L2 when compared to when they encounter the same dilemmas in their native language (L1). This suggests that language can significantly shape our emotional and cognitive responses to moral situations, influencing our choices and beliefs.

Given this complex interplay of culture, language and values in making moral judgments, one might ask: Do LLMs also exhibit a Foreign Language effect, changing their behavior when confronted with moral dilemmas in different languages? In this work, we extend the study by \citet{rao2023ethical} to five languages other than English, namely Spanish, Russian, Chinese, Hindi, Arabic, and Swahili. We probe three popular LLMs: GPT-4 \cite{openai2023gpt4}, ChatGPT (September 2023) \cite{chatgpt} and Llama2-70B-Chat \cite{touvron2023llama} systematically to assess their ethical reasoning abilities in different languages by prompting them to resolve an ethical dilemma reflecting conflicts between interpersonal, professional, social and cultural values, and a set of {\em ethical policies} (i.e., moral stances) that can help arrive at a clear resolution of the dilemma. These policies are drawn from three branches of normative ethics: \textbf{deontology} \cite{sep-ethics-deontological}, \textbf{virtue} \cite{sep-ethics-virtue} and \textbf{consequentialism} \cite{sep-consequentialism} and have three different levels of abstractions (similar to \citet{rao2023ethical}).   

Our study clearly demonstrates that indeed the LLMs exhibit different biases while resolving the moral dilemmas in different languages. This bias is minimal for English, in the sense that the resolution of the dilemma depends on the ethical policy rather than the model's own judgment, and it is maximum for low-resource languages such as Hindi and Swahili. This observation can also be interpreted as a reduced ethical reasoning capability of the LLMs in languages beyond English. However, as we shall see in this paper, the ethical reasoning ability (or conversely the bias) is dilemma-specific, which makes us conclude that the LLMs have strong value alignment biases in languages beyond English. Other salient findings are: (1) Across all languages, GPT-4 has the highest ethical reasoning ability, while Llama2-70B-Chat has the poorest; (2) across all models, the reasoning is poorest for Hindi and Swahili, while best for unsurprisingly, English, and also Russian; and (3) across all models, English and Spanish, and Hindi and Chinese have similar bias patterns.



%% file: related_works.tex
\section{Background}
The topic of right and wrong has been a subject of ongoing discussion among philosophers, psychologists, and other social scientists. Each field brings its unique perspectives and worries into this debate. In this section, we take these concerns as a guide to offer an overview of this ongoing discussion and how it connects with the field of machine ethics. Our main focus is on how these conversations affect the Natural Language Processing (NLP) community, and we also explore how Large Language Models (LLMs) can advance the frontiers of machine ethics, particularly in multilingual contexts.

\subsection{Ethics and Moral Philosophy}
{\em Ethics} is the branch of philosophy concerned with determining what is morally good or bad and what is considered right or wrong. It encompasses systems or theories of moral values and principles  \cite{kantlectureethics,kant1996metaphysics}. Within the realm of ethics, normative ethics plays a central role by seeking to establish standards of conduct for human actions, institutions, and ways of life. {\em Normative ethics} branches into {\em deontology}, which evaluates the inherent rightness or wrongness of actions based on moral rules or duties ~\cite{sep-ethics-deontological}; {\em virtue ethics}, which focuses on an individual's character and virtues rather than specific rules or consequences ~\cite{sep-ethics-virtue}; and {\em consequentialism}, which emphasizes the goodness or value of the outcomes or goals of actions~\cite{sep-consequentialism}.

{\em Ethical dilemmas} are situations characterized by conflicts between two or more moral values or principles, posing challenges for moral judgment and decision-making ~\cite{moral_dilemma}. The question of whether moral dilemmas can coexist with a consistent system of moral values is a subject of debate. Philosopher Williams argues that ethical consistency doesn't eliminate the possibility of moral dilemmas, as some actions that ought to be done may be incompatible \cite{williams1988ethical}. Resolving these dilemmas often requires making new value judgments within the existing ethical framework. 

{\em Value pluralism} is a key component of ethical dilemmas, suggesting that there are multiple values that can be equally correct and yet in conflict with each other \cite{james_1891_the_moral}. Different individuals or cultures may prioritize these values differently, resulting in varying resolutions of ethical dilemmas, each of which is ethically sound and consistent. Within the realm of pluralism, there exist various sub-schools of thought, including Rossian Pluralism \cite{the_right_and_the_good} and Particularism \cite{hare1965freedom}, each offering distinct viewpoints. Rossian pluralists advocate for the evaluation of moral principles based on their merits and demerits. Conversely, particularists contend that the assessment of moral pros and cons should be context-dependent. Nevertheless, both schools of thought share a fundamental conviction that there is no one-size-fits-all principle capable of resolving all moral conflicts. They also reject the notion of a rigid hierarchy of moral principles that could facilitate such resolutions. This perspective implies that there is no universally applicable set of moral values or principles that can address all situations and apply uniformly to all individuals.

\citet{inglehart2010wvs} introduced a framework for mapping global cultures which employs a two-dimensional axis system, where the x-axis represents a spectrum that stretches from survival ethics on the left to self-expression on the right and y-axis covers a range from tradition-based or ethnocentric moral views at the bottom to democratic and rational principles at the top. This visual representation illustrates the tendency of societies to move diagonally from the lower-left corner to the upper-right corner as they progress through industrialization and development.

\subsection{Foreign Language Effect on Morality}
Recent studies such as \cite{costa2014your, hayakawa2017thinking, corey2017our} have found a fascinating link between moral judgment and what's known as the "Foreign-Language Effect." This effect shows that people tend to make more practical choices when they face moral dilemmas in a foreign language (L2) compared to their native language (L1). This shift seems to be because using a foreign language makes people less emotionally connected to the situation, which, in turn, reduces the influence of emotions on their moral decisions. \citet{vcavar2018moral} also highlights that being better at and more comfortable with the foreign language (L2) can decrease this tendency to make practical decisions. This means that the language you use can significantly affect how you make moral choices, impacting many people. Furthermore, bilingual individuals' moral decision-making process is quite complex, involving factors like the type of dilemma, emotional excitement, and the language they're using \cite{chan2016effects}.

\subsection{Ethics in LLMs}
In the field of Ethics in NLP, most approaches assume a deontological perspective, with developers setting moral rules, but these may not readily apply to various contexts or Large Language Models (LLMs) ~\cite{onthemachinelearningofethicaljudgementsfromnaturallanguage}. \citet{awad_etal_2022_computational} introduce the Computational Reflective Equilibrium (CRE) framework for AI-based ethics, emphasizing moral intuitions and principles. \citet{sambasivan2021re}, \citet{bhatt-etal-2022-contextualizing} and \citet{ramesh-etal-2023-fairness} have raised questions of value-pluralism in AI and the need for recontextualizing fairness and AI ethics, particularly in global contexts. \cite{diddee_etal_2022_the} explore ethical concerns in Language Technologies for social good, emphasizing stakeholder interactions and strategies. \citet{choudhury2021linguistically} advocates the Rawlsian principle over utilitarianism in multilingual LLMs for linguistic fairness.

AI alignment seeks to ensure that AI systems conform to human goals and ethical standards, as highlighted by \cite{vox2018}. Various initiatives have put forth ethical frameworks, guidelines, and datasets to train and evaluate Language Models (LLMs) in terms of ethical considerations and societal norms \cite{hendrycks2020aligning, zhou2023rethinking, jiang2021can, rao2023ethical, tanmay2023exploring}. Moreover, \citet{tanmay2023exploring} introduces an ethical framework utilizing the Defining Issues Test to assess the ethical reasoning abilities of LLMs. However, it's worth noting that these efforts may be susceptible to biases based on the backgrounds of those providing annotations, as pointed out by \citet{olteanu2019social}. Recent research has placed a growing emphasis on in-context learning and supervised tuning to align LLMs with ethical principles, as demonstrated by the studies conducted by \citet{hendrycks2020aligning, zhou2023rethinking, jiang2021can, rao2023ethical, sorensen2023value}. These methodologies aim to accommodate a range of ethical perspectives, recognizing the multifaceted nature of ethics. \citet{rao2023ethical} posits that the generic ethical reasoning abilities can be infused into the LLMs so that they can handle value pluralism at a large scale. However, the authors have considered the ethical policies only in English.

In this study, we will be extending the work of \citet{rao2023ethical} with different languages to explore how LLMs behave when the multilingual ethical policies are infused into these LLMs . As far as we know, this study is the first of it's kind dealing with the ethics of LLMs in multilingual settings.

\subsection{Multilingual Performance of LLMs}
Language Models (LLMs) exhibit remarkable multilingual capabilities in natural language processing tasks, although their proficiency varies among languages \cite{zhao2023survey}. Their primary training data is in English, but they also incorporate data from various other languages, contributing to their generalisability \cite{brown2020language, chowdhery2022palm, zhang2022opt, zeng2022glm}. Nevertheless, significant challenges arise when LLMs interact with non-English languages, especially in low-resource contexts \cite{bang2023multitask, jiao2023chatgpt, hendy2023good, zhu2023multilingual}. Several studies suggest that enhancing their multilingual performance is possible through in-context learning and the strategic design of prompts \cite{huang2023not, nguyen2023democratizing}. Experiments conducted by \citet{ahuja2023mega} and \citet{wang2023seaeval} have brought to light an interesting aspect. They benchmarked LLMs across a range of Natural Language Processing (NLP) tasks, including Machine Translation, Natural Language Inference, Sentiment Analysis, Text Summarization, Named Entity Recognition, and Natural Language Generation. The results indicate that, while LLMs excel for a few well-resourced languages, they generally under-perform for most languages. Furthermore, \citet{kovavc2023large} have shown that LLMs exhibit context-dependent values and personality traits that can vary across different perspectives. This is in contrast to humans, who typically maintain more consistent values and traits across various contexts. 

Importantly, the current body of research has primarily concentrated on the technical capabilities of multilingual LLMs. There has been a relative lack of exploration into their moral reasoning within diverse linguistic and cultural contexts. Recognizing the significant impact of LLMs on real-world applications and domains, there is a growing need to delve into the ethical dimensions surrounding these multilingual language models.

%% file: results.tex
\section{Results and Observations}

Table~\ref{t:baselines} shows the baseline performance of all three models across various languages. GPT-4 resolves dilemmmas with a remarkably high agreement across all cases, with one notable exception observed in the Heinz dilemma when probed in Hindi, resulting in a 50\% agreement rate. An intriguing observation emerges when analyzing the GPT-4 results for the Rajesh dilemma; a distinct pattern of opposite resolutions (highlighted in red) is observed for most languages when compared to the English context (highlighted in green). Furthermore, ChatGPT exhibits conflicting behaviors between English and Hindi for all dilemmas in its resolution process. Llama2-70B-Chat exhibits the least consistent behavior among the three models, primarily due to the lower degree of agreement among all possible permutations of choices. Notably, Llama2-70B-Chat demonstrates a remarkable departure from the behavior of ChatGPT and GPT-4 in most dilemmas and languages. Llama2-70B-Chat tends to opt for affirmative resolutions, such as "should share" and "should steal the drug," more frequently, showcasing a distinct and contrasting behavioral pattern when compared to its counterparts. These findings hint at the potential presence of bias within the models towards specific dilemmas (and consequently, while resolving certain kinds of value conflicts) in specific languages, warranting further investigation.

Figure~\ref{fig:policy-based} provides a comprehensive overview of the results obtained from policy-based resolution by the models across various languages, comparing them to the ground-truth resolutions. GPT-4 consistently demonstrates superior ethical reasoning abilities across most languages, with the notable exception of Hindi. In stark contrast, Llama2-70B-Chat exhibits the least ethical reasoning capability across the board.

Table~\ref{t:mainresults} lists the accuracy of each model across the different levels of abstraction of the policies. We can see the trend that models tend to perform slightly better on average on lower abstraction levels. 

When considering the different policies, it becomes evident from Figure~\ref{fig:policy-based} that Level 2 policies, aligned with the consequentialist framework, are where the models predominantly excel, except for ChatGPT for Russian. For Level 1 deontological policies, Llama2-70B-Chat and ChatGPT perform well. These models perform well for Level 1 policies in virtue ethics except for Russian and Spanish. This highlights the nuanced interaction between policy levels, ethical frameworks, and model performance.

 GPT-4 exhibits improved reasoning ability when Level 2 policies are applied in Arabic, Russian, and Spanish, as compared to English within the consequentialist framework. Deontological policies in Level 2 work better for Russian and Spanish than for English. However, for virtue ethics-based policies, GPT-4 shows a distinct advantage for English. Conversely, Llama2-70B-Chat demonstrates notably superior performance with all ethical policies when expressed in English, as compared to all other languages. Overall, a general trend emerges where all models tend to perform less effectively in Hindi and Swahili, while achieving their best results in English and Russian. These observations provide valuable insights into the models' ethical reasoning abilities in various linguistic and ethical contexts.

Figure~\ref{fig:bias-confusion} provides a visual representation of our comparative analysis, illustrating the bias and confusion scores of each model across dilemmas and languages. From the figure, it becomes evident that GPT-4 exhibits the lowest levels of bias among the models, while ChatGPT demonstrates the highest bias levels. Interestingly, all models exhibit similar bias scores in both the English and Spanish, as well as in the case of Hindi and Chinese.


On comparing the confusion scores, Llama2-70B-Chat consistently shows the highest scores among all models, with GPT-4 displaying the lowest confusion scores. Notably, GPT-4's behavior varies across languages in the Rajesh dilemma, highest confusion score being observed for Swahili. In the context of Hindi, all models exhibit significantly divergent behavior when compared to their performance in English.